\documentclass[11pt,letterpaper]{article}
\usepackage[T1]{fontenc}
\usepackage{lmodern}
\usepackage[margin=1in]{geometry}
\usepackage{booktabs}
\usepackage{tabularx}
\usepackage{amsmath}
\usepackage{graphicx}
\usepackage[numbers,sort&compress]{natbib}
\usepackage{microtype}
\usepackage{caption}
\usepackage{comment}
\usepackage{etoolbox}
\usepackage{xurl}
\usepackage{setspace}
\usepackage[hidelinks]{hyperref}
\AtBeginEnvironment{table}{\small}
\excludecomment{CCSXML}
\newcommand{\ccsdesc}[2][]{\noindent #2\par}
\newcommand{\keywords}[1]{\smallskip\noindent\textbf{Keywords: }#1\par}
\newcommand{\Description}[2][]{}
\author{Siyi Yu\\Department of Marketing, Daniels School of Business\\Purdue University\\\texttt{yu1344@purdue.edu}}
\date{}
\title{Dull, Dirty, Dangerous? Working Conditions and the Limits of Occupational AI Exposure}
\hypersetup{pdftitle={Dull, Dirty, Dangerous? Working Conditions and the Limits of Occupational AI Exposure},pdfauthor={Siyi Yu}}

\begin{document}
\maketitle

\begin{abstract}
Automation is often justified as relief from dull, dirty, and dangerous work. I examine
how these conditions relate to occupational AI exposure using 21 O*NET items
across 754 US occupations. Higher exposure is associated with lower condition
scores under a large language model (LLM) measure and AI Occupational Exposure (AIOE), with equal occupation
weights and employment weights. Adjustment for physical demands, required education,
training and experience, and wages reduces the composite association. The adjusted
confidence intervals include zero for LLM exposure and remain below zero for AIOE.
Separating repetition from restricted discretion also changes what the combined Dull
score conveys. These comparisons show how item groupings and occupational characteristics
shape the interpretation of exposure estimates. Occupational profiles
offer a starting point for selecting contrasting research settings. Establishing desired
automation or realized relief requires evidence about workers' preferences and specific
interventions.
\end{abstract}

\begingroup\small
\noindent\textbf{CCS Concepts:}\par
\endgroup

\section{Introduction}

Promises to relieve dull, dirty, and dangerous work give automation a compelling purpose.
For HCI researchers, however, that purpose raises a concrete design question: which part
of a working day would a proposed system improve? A tool might shorten a task while leaving
noise, hazardous equipment, or restrictions on decision-making unchanged. To evaluate a
claim of relief, researchers need to connect the system's operation to the condition it is
intended to change. Occupational AI exposure scores describe technical potential across
jobs, but they do not make that connection on their own.

The distinction matters at two stages of research. When selecting a setting, researchers
need to understand the conditions associated with an occupation rather than infer them
from its exposure score. When evaluating a system, they need to establish whether the
proposed change addresses a condition workers want changed. Critical HCI and HRI research
has already challenged the assumption that technical automation necessarily improves work
\citep{rosie2025,laborprocess2026,ddd2026}. Studies of assembly workers, desired AI delegation,
and meaningful work likewise show why workers' assessments belong in that evaluation
\citep{welfare2019,shao2025,ranjit2026}. The question here concerns the occupational evidence
that can inform the first stage.

I examine how selected physical and organizational working conditions are distributed
across occupations relative to two published AI exposure measures. O*NET Work Context
provides ratings of occupational hazards, environments, repetitive activity, and discretion.
These ratings permit comparisons across occupations, including settings outside the
computer-assisted tasks emphasized in some recent HCI research. Their breadth comes at a
cost: an occupational average cannot describe every workplace or distinguish the conditions
different workers value, tolerate, or want changed. I therefore use the ratings to describe
working conditions, reserving questions of desirability for evidence about workers' own
judgments.

Existing research already relates AI exposure to job quality and working conditions.
\citet{parteka2024} compare working conditions across occupations exposed to AI, software,
and robots. \citet{pizzinelli2023} combine O*NET work contexts and occupational preparation
with AIOE to examine potential complementarity. \citet{giuntella2025} examine AI exposure
alongside wellbeing and physical job intensity in German longitudinal data. These studies
make the broad connection between AI and conditions at work an established research topic.
The contribution of another occupational comparison depends on what its measures reveal
and how its conclusions change under plausible alternative specifications.

My analysis focuses on those choices. I organize 21 Work Context items into
domains associated with the Dull, Dirty, and Dangerous (DDD) formulation and link them to
LLM exposure \citep{eloundou2024} and AI Occupational Exposure (AIOE) \citep{felten2021}.
Several selected items describe physical work settings, whereas contemporary AI exposure
is substantial in cognitive work. A negative association between the condition scores
and AI exposure may therefore reflect familiar differences in occupational task content.
I examine how that association changes after accounting for physical demands, occupational
preparation---the education, training, and experience a job requires---and wages.

The Dull domain raises a second question about interpretation. Repeating movements or
working at a machine's pace differs from having little authority to determine tasks and
priorities. Combining these conditions may be convenient for summarizing occupations, but
it can obscure which condition is associated with AI exposure. I report the
composite and compare its results with a separate analysis of repetition and machine pacing,
and restricted discretion. The analysis addresses two research questions:

\begin{itemize}
\item \textbf{RQ1:} How do the selected working conditions relate to LLM exposure and AIOE
under equal occupation weights and employment weights?
\item \textbf{RQ2:} How do these associations change when occupational characteristics are
included in the models and when the condition index is constructed differently?
\end{itemize}

The primary models associate higher AI exposure with lower scores in every domain and in
the composite. Adjustment substantially reduces the composite association: the fully
adjusted LLM intervals include zero, while the AIOE intervals remain below zero. Within
Dull, the LLM estimates are more negative for restricted discretion than for repetition and
machine pacing, although I do not test the difference between those coefficients. These
findings distinguish the direction of an aggregate association from the information its
components and adjusted models provide.

The paper contributes a comparison of working-condition and AI exposure
measures, together with evidence about how their interpretation depends on item grouping
and occupational characteristics. It also describes occupational profiles that keep
separate condition scores. A worked comparison shows how occupations can enter the same
high-score, low-exposure group through different domains. Such profiles can help researchers
formulate questions for contrasting settings; their usefulness for recruitment or design
has not been evaluated here. All analyses are observational, and the adjustment and
decomposition analyses are exploratory. This account keeps the occupational patterns
separate from the further evidence needed to establish desired automation or relief.

\section{Related Work}

\subsection{AI exposure and conditions at work}

Task-based studies of technological change distinguish cognitive from manual work and
routine from nonroutine tasks \citep{autor2003}. These distinctions help explain why
different technologies reach different occupations. They also provide a reference point
for interpreting an association between AI exposure and working conditions: a new condition
index may partly summarize task differences that existing occupational classifications
already capture. O*NET-based teleworkability research illustrates this overlap. Outdoor work,
protective equipment, and other physical or on-site requirements help distinguish jobs that
can be performed at home \citep{dingel2020}. Those features also appear in the condition
items examined here.

The two primary exposure measures describe different forms of technical potential.
The LLM measure assesses whether a language model, used alone or with additional software,
could reduce the time needed for occupational tasks \citep{eloundou2024}. AIOE relates AI
applications to the abilities required in an occupation \citep{felten2021}. Neither measure
observes deployment or job loss. Both use O*NET information, so their agreement cannot be
treated as agreement between wholly independent sources. Comparing them nevertheless
reveals whether a condition association persists across their different technological
scopes and constructions.

Close labor research already connects these questions to working conditions. Table~\ref{tab:position} summarizes the closest comparisons.
\citet{pizzinelli2023} use O*NET work contexts and Job Zones to characterize potential
complementarity alongside AIOE. Their question concerns circumstances under which AI may
complement or substitute for labor. My index instead summarizes selected hazards,
environmental conditions, repetition, and restricted discretion. Thus the distinction
rests on the constructs and comparisons being examined, rather than on combining O*NET
conditions with AI exposure for the first time.

\citet{parteka2024} compare AI, software, and robot exposure in relation to multidimensional
working conditions in Europe. Their results place relatively favorable conditions in
AI-exposed occupations within a broader comparison of technologies. \citet{giuntella2025}
study wellbeing, health, and physical job intensity using German longitudinal data.
My cross-sectional US linkage does not estimate changes in those outcomes. It examines
how associations with selected condition scores depend on the exposure measure, physical
demands and other occupational characteristics, and the grouping of repetition with
restricted discretion. This narrower comparison complements research on job quality and
worker outcomes without substituting for it.

\subsection{Worker preferences, meaning, and control}

Occupational conditions and workers' judgments answer different questions.
\citet{takayama2008} related public judgments of robot suitability to O*NET occupations,
establishing an early occupation-wide comparison. Public judgments, however, differ from
the preferences of people doing the work. \citet{welfare2019} found that assembly workers
valued aspects of physical and social work while seeking relief from safety problems and
rushing. \citet{shao2025} compare workers' desired AI delegation with assessed agent
capabilities and desired human agency. These studies supply direct evidence about valued
work and desired assistance that occupational condition scores cannot provide.

\citet{ranjit2026} bring worker valuation and AI exposure particularly close together.
They compare meaningful work and busywork with exposure using human task ratings and
ratings scaled with language models, concentrating on computer-assisted tasks. My analysis
focuses on physical environments, hazards, repetition, and discretion across occupations.
The studies consequently describe different parts of the work experience. A physically
demanding activity can be meaningful; an activity with high technical exposure can still
be one that a worker wishes to retain. Neither meaningfulness nor a condition score alone
establishes an overall judgment about a job.

Worker-centered HCI also examines who decides what changes. \citet{rosie2025} identify
automation arguments that can disempower workers, including assumptions about the
desirability of their jobs. Labor process theory directs attention to managerial control,
consent, resistance, and value production \citep{laborprocess2026}. These perspectives
explain why a faster task does not necessarily give workers more time or autonomy.
They motivate keeping a technical exposure estimate separate from the organizational
arrangement in which a system would be used.

Deployment can also create work outside the task being automated. Studies of AI integration
in essential work describe calibration, troubleshooting, and repair \citep{fox2023patchwork}.
Participatory research in hospitality identifies worker-generated changes to algorithmic
room assignment concerning workload and control \citep{spektor2023}. Low LLM exposure or
AIOE therefore does not describe the full technological environment of an occupation.
Algorithmic management, robotics, and other forms of automation may matter even where the
capabilities represented by these particular indices have limited reach.

\subsection{What DDD-related condition scores represent}

The DDD formulation brings together conditions with different meanings.
\citet{ddd2026} examine its use in robotics and argue for interpretation grounded in work
settings and workers' sense-making. Physical contamination differs from social or moral
taint; repetitive motion differs from experienced boredom. I use DDD as an organizing
vocabulary for selected items, rather than assume that the label identifies a single
trait or a complete taxonomy of burdensome work.

Research on dirty work makes the distinction between conditions and valuation explicit.
Occupational stigma can coexist with positive identity \citep{ashforth1999}, and workers
in physically dirty jobs can report satisfaction and self-worth \citep{deery2019}.
Work-design research likewise treats decision latitude as more than an aspect of dullness
\citep{karasek1979}. Demands and resources can have different relationships with exhaustion
and disengagement \citep{demerouti2001}. A score that combines repetition with restricted
discretion should therefore make that combination visible instead of letting the domain
name imply a shared subjective experience.

The condition index covers only part of job quality. Broader frameworks include earnings,
security, and the working environment \citep{cazes2015}; the selected items omit many
psychosocial aspects of the last category. I keep wages outside the index to distinguish
conditions from compensation. A high score describes more of the selected conditions as
coded, not how a worker weighs those conditions against pay, meaning, security, or other
features of a job.

This interpretation also determines the role of measurement checks. The index is a
content-defined summary: its meaning comes from the selected items and their weights.
It is not treated as a reflective scale in which a single underlying trait causes every
item response \citep{bollen1991,streiner2003,jarvis2003}. Correlations and internal-consistency
coefficients describe how the items vary together, but they cannot establish whether
important conditions have been omitted. The item definitions and sensitivity comparisons
are therefore central to interpreting the results.

\begin{table}[t]
\caption{Closest empirical comparisons. Differences in constructs and research questions define the present study's scope.}
\label{tab:position}
\begin{tabularx}{\linewidth}{l>{\raggedright\arraybackslash}X>{\raggedright\arraybackslash}X}
\toprule
Study & Evidence and question & Relation to this study \\
\midrule
\citet{pizzinelli2023} & O*NET contexts, Job Zones, and AIOE; potential complementarity & Shares occupational inputs; measures complementarity rather than the selected condition burdens. \\
\citet{parteka2024} & Technology exposure and multidimensional working conditions in Europe & Establishes a close precedent for the broad condition--exposure association. \\
\citet{giuntella2025} & German longitudinal data; AI exposure, wellbeing, health, and physical job intensity & Examines worker outcomes over time; the present linkage compares occupational scores. \\
\citet{ranjit2026} & Task-level meaning, busywork, and AI exposure & Examines valuation of computer-assisted tasks; this study examines physical and organizational conditions. \\
\citet{welfare2019,shao2025} & Workers' valued activities and desired assistance or delegation & Provides preference evidence beyond the scope of occupational condition ratings. \\
Present study & Selected condition scores, two AI exposure measures, and specification comparisons & Examines adjustment, item grouping, and the interpretation of occupational profiles. \\
\bottomrule
\end{tabularx}
\end{table}

\section{Data and Method}

\subsection{Data sources and unit of analysis}

The unit of analysis is a six-digit US Standard Occupational Classification (SOC)
occupation. I linked O*NET 31.0 Work Context ratings and Job Zones to the published AI
exposure measures and public Bureau of Labor Statistics (BLS) data. The main analyses use
employment and wages information from the Occupational Employment and Wage Statistics (OEWS) program. Supplementary
analyses use workforce composition from the Current Population Survey (CPS), fatality
counts from the Census of Fatal Occupational Injuries (CFOI), and a hybrid exposure measure.
Table~\ref{tab:sources} gives the source versions and observation periods.

\begin{table}[t]
\caption{Data sources and observation periods. Employment and workforce composition come from BLS files distributed in the public Eloundou et al.\ repository (\url{https://github.com/openai/GPTs-are-GPTs}). BLS publishes the CFOI tables at \url{https://www.bls.gov/iif/fatal-injuries-tables.htm}.}
\label{tab:sources}
\begin{tabularx}{\linewidth}{l>{\raggedright\arraybackslash}X>{\raggedright\arraybackslash}X}
\toprule
Source & Version and unit & Fields or use \\
\midrule
O*NET & 31.0, August 2026 release; detailed O*NET-SOC & Work Context ratings, suppression flags, Job Zones, task statements, SOC crosswalk \citep{onet31} \\
Eloundou et al. & Published exposure data; detailed O*NET-SOC & GPT-4-rated \texttt{dv\_rating\_alpha}, \texttt{beta}, \texttt{gamma} columns in \texttt{occ\_level.csv} \citep{eloundou2024} \\
AIOE & 2021 dataset, SOC 2010 & Appendix A of \path{AIOE_DataAppendix.xlsx} \citep{felten2021} \\
BLS OEWS & May 2021; detailed occupation & \texttt{TOT\_EMP}, \texttt{A\_MEDIAN}; employment weights and wages \citep{oewsfaq} \\
BLS CPS & 2022 annual Table 11; occupation category & Sex, race, and Hispanic or Latino composition; annual workforce shares \\
BLS CFOI & 2024 Table A-5; occupation title & Fatality counts for an exploratory measurement check \\
Anthropic & March 2026 release; August/November 2025 traffic & \path{labor_market_impacts/task_penetration.csv}; hybrid exposure scores \citep{massenkoff2026,massenkoff2026appendix} \\
\bottomrule
\end{tabularx}
\end{table}

\subsection{Occupational linkage and samples}

I retained Work Context ratings on the source's CX scale unless they were flagged
``Recommend Suppress = Y.'' These ratings use item-specific response categories, described
below. I averaged ratings across detailed O*NET occupations sharing a six-digit SOC code.
Detailed specialties received equal weight because corresponding employment weights were
unavailable. O*NET contributed 911 detailed occupations with usable
ratings, aggregated into 786 six-digit occupations. Suppression can make the
specialties contributing to an occupation's average differ from item to item.

I also averaged the Eloundou exposure scores within six-digit codes. AIOE required a
crosswalk from SOC 2010 to the SOC 2018-based classification used by O*NET. I used the
official crosswalk, removed duplicate source--target pairs, and averaged unique source
values mapped to each target. Direct-code matching was used only when no crosswalked
value was available. Appendix~\ref{app:linkage} gives the exact files and processing details.

The reference set with condition items and both AI exposure measures contained
781 occupations. Loan Interviewers and Clerks had no observed item
in the Dangerous domain after suppression and could not receive a composite score.
Excluding this occupation left 780 scored occupations. A further
26 lacked usable positive employment, leaving the primary sample of
754 occupations in 22 major groups. This sample represented
129,550,700 jobs, or 92.0\% of national detailed
OEWS employment. Models with equal occupation weights and employment weights used the
same primary sample. An additional analysis with equal occupation weights retained all
scored occupations.

The adjustment models required complete physical-demand, Job Zone, wage, and employment
data. I held this smaller sample constant across all adjustment stages, so changes in the
coefficient did not reflect changes in which occupations entered the model.
Table~\ref{tab:samples} distinguishes this sample from those used for supplementary
composition, fatality, and hybrid-exposure analyses. The supplementary samples answer
different questions and should not be read as extensions of a single common population.

\begin{table}[t]
\caption{Samples used in each analysis. Employment coverage is relative to the primary sample unless labeled national.}
\label{tab:samples}
\begin{tabularx}{\linewidth}{Xrr}
\toprule
Analysis & Occupations & Employment coverage \\
\midrule
Condition scores and both AI exposure measures & 780 & --- \\
Primary comparison, both weighting schemes & 754 & 92.0\% national \\
Common sample for sequential adjustment & 736 & --- \\
CPS title matches with complete composition data & 244 & 72.1\% \\
Fatality-ratio diagnostic & 170 & --- \\
Hybrid exposure linkage & 753 & 99.7\% \\
\bottomrule
\end{tabularx}
\end{table}

\subsection{Working-condition index}

The index uses 21 items in three domains.
Dangerous included six hazard items and two items about protective equipment. Dirty
included outdoor work and adverse environmental conditions. Dull included repetitive
motions, repeated tasks, machine pacing, and two items about discretion. The latter items
were reversed so that higher values denoted less discretion. Appendix Table~\ref{tab:items}
lists every item and identifier.

I refer to an occupation's resulting value as its condition score. ``Measured burden'' is
shorthand for that selected and oriented content. It does not imply that every item is
uniformly unwanted: protective equipment and outdoor work, for example, can have different
meanings across settings. The domain names identify the item groupings used in this
analysis; descriptions of particular conditions remain necessary when interpreting them.

Work Context ratings use item-specific anchors on a one-to-five scale. I reversed the
discretion ratings as $6-x$, standardized each item, and averaged the observed items within
each domain. I then standardized the domain averages and averaged the three domains to
form a final standardized composite. This gives the domains equal weight, so each Dull
item contributes more than each item in either of the larger physical domains.

Two reference populations enter the calculation. Let $R$ be the exposure-linked reference
set, and let $S$ be the scored sample obtained by excluding occupations with an entirely
undefined domain. Let $J_{id}$ contain the observed items for occupation $i$ in domain $d$,
and let $\mathcal{Z}_{A}$ denote centering and division by the population standard deviation
in population $A$. For an oriented item rating $x_{ij}$, the calculation is
\begin{equation}
z_{ij}=\mathcal{Z}_{R}(x_{ij}),\qquad
D_{id}=\mathcal{Z}_{S}\!\left(\frac{1}{|J_{id}|}\sum_{j\in J_{id}}z_{ij}\right),\qquad
B_i=\mathcal{Z}_{S}\!\left(\frac{D_{i,\mathrm{Dull}}+D_{i,\mathrm{Dirty}}+D_{i,\mathrm{Dangerous}}}{3}\right).
\end{equation}
All standard deviations use the population divisor. Item standardization thus uses $R$,
whereas domain and composite standardization use $S$. These reference populations remain
the same when employment weights enter the regressions.

An occupation could receive a domain score if at least one item in that domain was observed.
There were 25 scored occupations with at least one missing selected
item and 755 with all items observed. This available-item rule can
change the effective item weights across occupations. I retained it in the primary score
and examined complete-item cases separately.

I used standardized-item coefficient alpha and domain correlations to describe covariance
among the selected conditions. I also examined correspondence with an occupational fatality
ratio, whose time periods and population coverage are imperfectly aligned. Appendix~\ref{app:diagnostics} reports alpha and the fatality-ratio
checks, and the Dirty--Dangerous correlation appears in the results. They do not provide worker-based content
validation, test--retest evidence, or a direct measure of experienced boredom. The index's
interpretation rests on its stated content and the consequences of alternative scoring
choices.

\subsection{AI exposure measures}

The primary LLM exposure measure was the GPT-4-rated beta series in the published Eloundou
dataset \citep{eloundou2024}. Its task rubric distinguishes exposure to a standalone model
from exposure requiring additional software. The occupational beta score combines those
assessments; it is not a record of workers completing tasks with AI. The source's human-rated
series was not used in the present comparisons. AIOE was taken from the 2021 ability-based
dataset \citep{felten2021} and harmonized as described above.

I standardized each measure with equal occupation weights across the scored sample and
entered them in separate models. This puts each coefficient in condition-score standard
deviations per AI-exposure standard deviation. The standardization facilitates comparison
of reported associations, but does not make the measures equivalent in technological scope.
I report both measures under both regression weighting schemes throughout the main
comparisons.

Two supplementary analyses examine other exposure constructions. One separates the LLM
score into standalone-model and additional-software components. The other uses a hybrid
score combining theoretical feasibility with classified platform activity and workflow
weighting \citep{massenkoff2026,massenkoff2026appendix}. Conversation-based research can
describe platform activity \citep{anthropic2025}, but the hybrid score is partly constructed
from theoretical exposure. It therefore cannot serve as independent evidence of population
adoption. Appendix~\ref{app:supplementary} reports both constructions and their results.

\subsection{Association models and adjustment}

For each standardized condition outcome $Y_i$, I estimated
\begin{equation}
Y_i=\mu_{g(i)}+bE_i+\epsilon_i,
\end{equation}
where $g(i)$ identifies the SOC major group and $E_i$ is one standardized AI exposure
measure. The major-group fixed effects compare occupations within broad occupational
families. I estimated the composite and each domain using equal occupation weights and
OEWS employment weights normalized to mean one. The former describe the occupational
distribution; the latter give more influence to occupations representing more jobs.
Neither scheme recovers within-occupation variation. I also estimated models without
major-group effects to describe the association across occupational families.

The adjustment analysis asks whether the condition--exposure association remains comparably
large after including physical demands and other occupational characteristics. I constructed
a separate physical-demand score from standardized climbing, kneeling/crouching,
bending/twisting, and standing ratings, then standardized its average. Occupational
preparation was represented by O*NET Job Zones, which combine required education, training,
and experience. Wage was the log of the occupational median annual wage. These variables
provide a specified occupational comparison, rather than an exhaustive representation of
physical or cognitive task content.

Every adjustment stage used the same complete sample. The baseline included AI exposure
and major-group fixed effects. The next stage added the physical-demand score as a
predictor with the same condition outcome. The final stage also included
Job Zone indicators and log median wage. I calculated exposure's partial $R^2$ as the
reduction in weighted residual sum of squares when exposure was added to the otherwise
identical model. This summarizes the additional variation in the condition score associated
with exposure under that specification.

The adjustment coefficients are descriptive. The included variables have not been
identified as causal confounders, and a smaller exposure coefficient does not show that
physical demands caused the baseline association. Nor does adjustment imply that physical
conditions should be discounted when evaluating workers' experiences. Its purpose is to
show how much the occupational comparison depends on characteristics already familiar
from research on task content.

\subsection{Uncertainty and specification checks}

I used cluster-robust CR1 covariance at the SOC major-group level and a Student $t$
reference distribution with $G-1$ degrees of freedom, where $G$ is the number of groups.
Intervals are two-sided 95\% confidence intervals. The modest number of groups makes
this an approximation \citep{cameron2015}; the intervals do not incorporate all uncertainty
in item selection, source ratings, or linkage. Appendix~\ref{app:linkage} gives the
finite-sample multiplier. Holm adjustment covers the sixteen tests in the primary
domain-by-measure-by-weighting comparison. All secondary intervals are exploratory and
fall outside that primary adjustment for multiple comparisons.

The specification checks address distinct choices. To examine index construction, I gave
all items equal weight, restricted scoring to complete-item cases, omitted protective-equipment
items, omitted outdoor items, and constructed an index with physical demands as a fourth
domain. In the last check, physical demands change the outcome itself; this differs from
including the physical-demand score as a predictor in the adjustment models. I also
estimated repetition and machine pacing separately from restricted discretion to examine
what the combined Dull score conceals.

To examine sample dependence, I repeated the primary composite models while omitting each
major group in turn and used all scored occupations in models with equal occupation weights.
These checks assess influential groups and employment-data exclusions, and
Appendix~\ref{app:diagnostics} summarizes them. I did not
select the primary items or exposure measure to maximize that association.

The study is observational and was not independently preregistered, and analyst blindness
was not verified. The adjustment models, repetition/discretion split, and other secondary
comparisons are exploratory.

\subsection{Occupational profiles}

To describe occupations with higher measured burdens and lower AI exposure, I selected
those above the composite median and at or below the medians of both exposure measures.
All medians gave equal weight to occupations in the primary sample. The contrast group
had a composite score at or below its median and both exposure scores above their medians.
I also reported classifications based on either exposure measure separately and a stricter
definition using the upper composite tertile and the lower tertiles of both exposure measures.
These cutoffs identify relative groups within the data; they are not thresholds of harm
or technical feasibility.

For each occupation, a profile consists of its three domain scores alongside its classification, employment,
and wage. Reading these scores side by side distinguishes the conditions behind a composite label.
I use the profiles below to illustrate how occupations in the same group can have different leading
domains. Supplementary workforce-composition comparisons use the same joint groups and
report their separate CPS linkage coverage in Appendix~\ref{app:composition}.

\section{Results}

\subsection{Higher AI exposure is associated with lower condition scores}

\begin{figure}[t]
\centering\includegraphics[width=\linewidth]{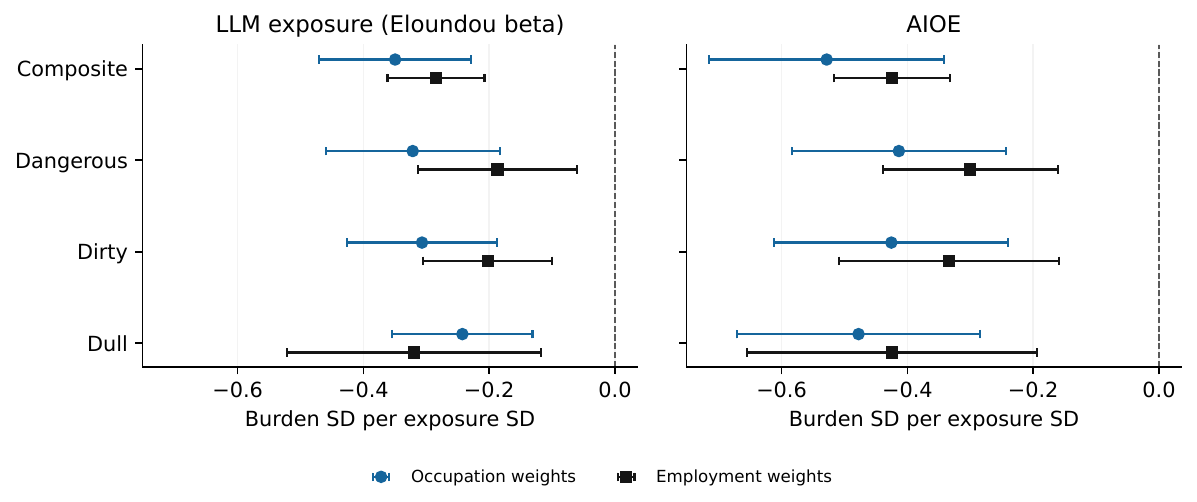}
\caption{AI exposure and working-condition scores in the primary sample. Points are regression coefficients, with 95\% confidence intervals based on CR1 covariance and a $t$ reference distribution. Every model includes major-group fixed effects. Circles denote equal occupation weights; squares denote employment weights. Exact values appear in Appendix Table~\ref{tab:primary}.}
\Description{The left panel shows LLM exposure and the right panel AIOE. Composite, Dangerous, Dirty, and combined Dull coefficients are negative under both weighting schemes. Employment weights bring most estimates closer to zero; the LLM coefficient for the combined Dull domain instead becomes more negative. All displayed intervals lie below zero.}
\label{fig:primary}
\end{figure}

Higher AI exposure is associated with lower composite and domain scores under both measures
and both weighting schemes (Figure~\ref{fig:primary}). The primary models compare occupations
within major groups on the common sample of 754 occupations. All primary
confidence intervals lie below zero; the largest Holm-adjusted $p$-value in this family is
0.007. Appendix Table~\ref{tab:primary} reports the complete estimates.

For the composite, the LLM coefficients are $-$0.350
with equal occupation weights and $-$0.285 with employment
weights. The AIOE coefficients are $-$0.528 with equal
occupation weights and $-$0.425 with employment weights.
Thus the composite coefficients are closer to zero when occupations with more jobs receive
more weight. The domain-level pattern is not identical: the LLM coefficient for the combined
Dull domain becomes more negative under employment weights.

Part of the overall association lies between broad occupational families. Under equal
occupation weights, the LLM composite coefficient changes from
$-$0.674 without major-group effects to
$-$0.350 with them. The corresponding AIOE coefficient
changes from $-$0.793 to
$-$0.528. Including major-group effects reduces the magnitude
for both measures while retaining a negative association within families.

The condition domains and exposure measures themselves overlap. Dirty and Dangerous
scores correlate at $r=0.804$. LLM exposure and AIOE
correlate at Pearson $r=0.833$ and Spearman
$\rho=0.853$ across the scored sample. Their agreement accompanies
differences in the adjusted and component results reported next. Further covariance and
fatality-ratio diagnostics appear in Appendix~\ref{app:diagnostics}.

\subsection{Physical-demand adjustment reduces the composite association}

\begin{figure}[t]
\centering\includegraphics[width=\linewidth]{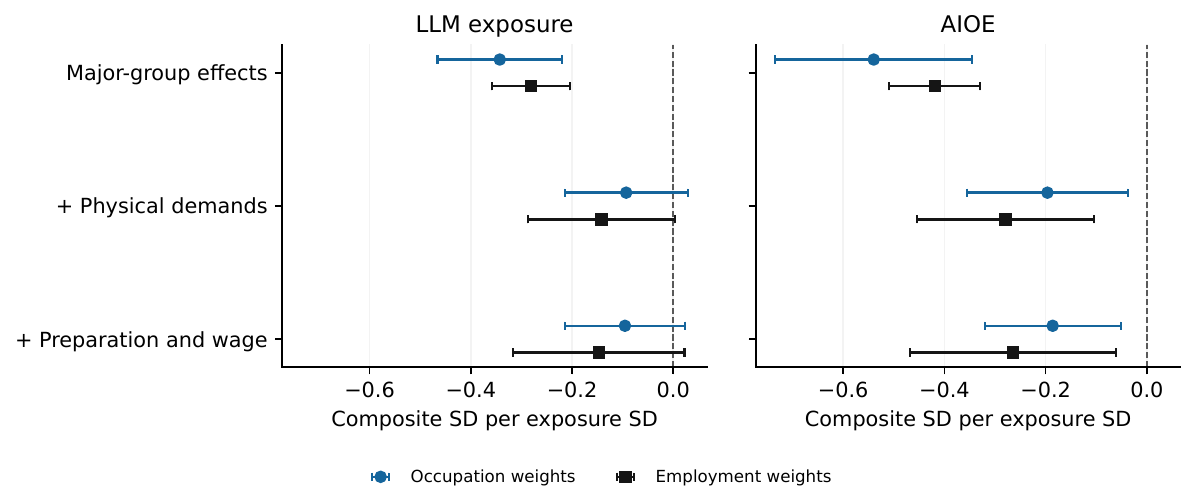}
\caption{Adjustment models on the same complete sample. Each stage retains major-group fixed effects. Physical demands enter as a predictor at the second stage; Job Zone indicators and log median wage enter at the third. Circles denote equal occupation weights and squares employment weights. Exploratory 95\% intervals use CR1 covariance and a $t$ reference distribution.}
\Description{The magnitude of the negative composite association falls for both AI exposure measures after physical demands enter the model. In the fully adjusted models, LLM confidence intervals include zero and AIOE intervals remain below zero under both weighting schemes. These intervals describe uncertainty within each model, rather than a formal comparison between the two measures.}
\label{fig:adjustment}
\end{figure}

The estimated composite association becomes much smaller when physical demands enter the
model (Figure~\ref{fig:adjustment}). Every stage uses the same
736 occupations. Under equal occupation weights, the LLM coefficient
changes from $-$0.343 in this sample's baseline
model to $-$0.093 after physical-demand adjustment.
Adding Job Zone indicators and log median wage yields
$-$0.095, with confidence interval
[$-$0.214, 0.023].

The fully adjusted LLM coefficient with employment weights is
$-$0.147, with interval
[$-$0.316, 0.022]. Both LLM intervals include zero.
The AIOE coefficient is $-$0.186
[$-$0.320, $-$0.052] under equal occupation weights and
$-$0.264
[$-$0.468, $-$0.060] under employment weights. Both AIOE intervals
remain below zero. These are separate within-model results; they do not constitute a
formal test that the coefficients differ between exposure measures.

Exposure accounts for a limited additional share of variation after adjustment. Its partial
$R^2$ is 1.1\% for LLM exposure and
2.4\% for AIOE under equal occupation weights.
With employment weights, the corresponding values are
3.4\% and
7.6\%.
The weaker adjusted LLM association qualifies the magnitude seen in the primary models;
its interval does not establish an exactly zero population association.

\subsection{Repetition and restricted discretion give different descriptive patterns}

\begin{figure}[t]
\centering\includegraphics[width=\linewidth]{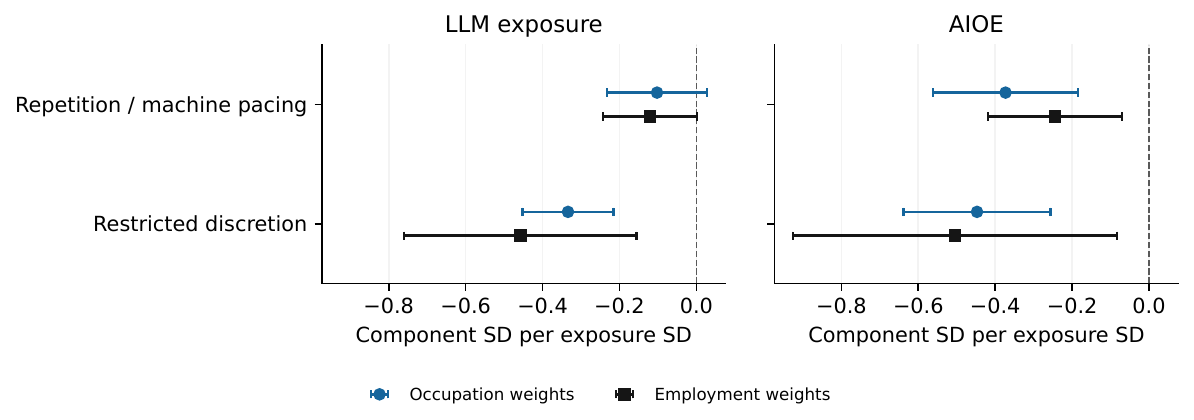}
\caption{The Dull domain separated into repetition and machine pacing, and restricted discretion. Each component is standardized separately. Circles denote equal occupation weights and squares employment weights; exploratory 95\% intervals use CR1 covariance and a $t$ reference distribution. Models retain major-group fixed effects.}
\Description{LLM estimates are more negative for restricted discretion than for repetition and machine pacing. The repetition intervals include zero under both weighting schemes, with the employment-weighted upper bound close to zero. Restricted-discretion intervals lie below zero. AIOE estimates and intervals are negative for both components and weighting schemes. No formal test of the difference between the component coefficients is reported.}
\label{fig:components}
\end{figure}

The LLM point estimates are more negative for restricted discretion than for repetition
and machine pacing (Figure~\ref{fig:components}). Under equal occupation weights, the
repetition coefficient is $-$0.102
[$-$0.233, 0.028], and the restricted-discretion coefficient is
$-$0.334
[$-$0.452, $-$0.216]. Under employment weights, the repetition
coefficient is $-$0.121
[$-$0.242, 0.001], and the restricted-discretion coefficient is
$-$0.458
[$-$0.759, $-$0.156].

The LLM repetition intervals include zero under both weighting schemes, with the
employment-weighted upper bound close to zero. The restricted-discretion intervals lie
below zero. AIOE intervals lie below zero for both components and both weighting schemes.
The two component scores correlate at $r=0.444$.
Reporting only their combined Dull score therefore conceals the separate estimates and
their uncertainty. This comparison is descriptive: no formal test of the difference
between the component coefficients was conducted.

The sign of the composite coefficient persists across several changes to index content.
Under equal occupation weights, the LLM coefficient is
$-$0.348 when items receive equal weight,
$-$0.340 when protective-equipment items are omitted, and
$-$0.343 when outdoor items are omitted. Composite
coefficients also remain negative in the complete-item analysis and when physical demands
form an additional index domain. These checks concern the direction of the composite
association under alternative scoring rules. They do not erase the differences in
component estimates or the attenuation after regression adjustment.

\subsection{Similar classifications contain different occupational profiles}

The joint median-based group contains 286 occupations representing
40,957,610 jobs, or 31.6\% of
employment in the primary sample. The median of the occupational median wages in this
group is \$45,720. The contrast group contains
286 occupations and 54,519,810 jobs.
Table~\ref{tab:sets} shows how the group changes when the cutoffs or exposure measures change.

\begin{table}[t]
\caption{Occupations with higher composite scores and lower AI exposure under alternative relative cutoffs. All definitions use the primary sample and the same employment denominator; cutoffs give equal weight to occupations.}
\label{tab:sets}
\begin{tabularx}{\linewidth}{Xrrr}
\toprule
Definition & Occupations & Jobs & Share \\
\midrule
High score, low LLM exposure & 300 & 41,422,280 & 32.0\% \\
High score, low AIOE & 315 & 47,356,980 & 36.6\% \\
High score, low on both (medians) & 286 & 40,957,610 & 31.6\% \\
High score, low on both (tertiles) & 157 & 20,164,070 & 15.6\% \\
\bottomrule
\end{tabularx}
\end{table}

Construction laborers, heavy truck drivers, nursing assistants, restaurant cooks,
landscaping workers, and housekeeping cleaners appear in the joint group. Their domain
profiles show why membership alone is an incomplete description. Construction laborers
have their highest standardized domain score in Dirty (2.07; Dangerous 0.91, Dull 0.89), whereas housekeeping cleaners
have their highest score in Dull (1.18). For housekeeping cleaners, both physical-domain scores
are below the scored-sample means (Dangerous $-$0.11, Dirty $-$0.78) despite membership in the higher-composite group.
The shared classification thus does not imply the same pattern of environmental conditions,
hazards, repetition, and restricted discretion.

These comparisons do not treat the composite as a priority ranking. An occupational profile identifies which measured conditions contribute
to a score; it does not identify the task causing those conditions or the intervention
workers would welcome. Appendix~\ref{app:composition} reports workforce composition within
the same joint groups, with the coverage limits of that additional linkage. Appendix~\ref{app:supplementary}
reports the exposure-component and hybrid-score results, including the hybrid estimate
whose employment-weighted interval includes zero.

\section{Discussion}

\subsection{Interpreting the association beyond its direction}

The primary models consistently place higher AI exposure alongside lower scores on the
selected conditions. This pattern fits prior evidence that contemporary AI exposure is
substantial in cognitive occupations and differs from exposure to earlier forms of
automation \citep{eloundou2024,felten2021,parteka2024}. The adjustment analysis specifies
how much of the present association remains when physical demands, preparation, and wages
enter the comparison. Much of the LLM coefficient's magnitude disappears after the
physical-demand score enters, and the fully adjusted interval includes zero under both
weighting schemes. AIOE retains intervals below zero, although its estimated association
also becomes smaller.

For an occupational condition index, this distinction changes the contribution one can
claim. Reproducing a negative association across several scoring variants establishes a
pattern within those specifications. It does not establish that the index adds substantial
information beyond familiar physical and cognitive differences between jobs. The small
partial $R^2$ values and attenuated coefficients make that question explicit. They also
show why reporting both exposure measures matters: a conclusion drawn from one set of
adjusted intervals would omit the other measure's result.

The adjustment is not a causal explanation. Physical demands can overlap with the
conditions summarized by the index, and the models leave other occupational characteristics
unmeasured. AIOE's remaining association could reflect occupational abilities, omitted
conditions, or shared measurement inputs. \citet{pizzinelli2023} use related inputs to
examine potential complementarity; the present models instead ask about associations with
selected conditions. The two questions should remain distinct even when they use the
same occupational database.

Claims about worker benefit face a separate evidential boundary. Neither a large primary
coefficient nor a persistent adjusted coefficient would supply the worker preference and
intervention-outcome measurements absent here. Research that observes wellbeing or health
over time, such as \citet{giuntella2025}, addresses outcomes this linkage does not observe.
The contribution of the present analysis is to make the occupational comparison more
specific, while leaving evaluation of actual changes in work to an appropriate design.

\subsection{Specify the condition behind a promise to reduce dull work}

The Dull comparison shows how a familiar label can compress different conditions into one
number. The LLM coefficients for restricted discretion are more negative than those for
repetition and machine pacing, and their intervals have different relationships to zero.
Although the analysis does not formally test that coefficient difference, it demonstrates
the information lost when only the combined estimate is reported. A negative Dull
coefficient should not be paraphrased as a result about every form of repetitive work.

This finding connects to research distinguishing meaningful work, busywork, and desired
delegation \citep{ranjit2026,shao2025}. Repetition can take different forms, and authority
over task choice is a separate feature of work. For example, repetitive information
handling and repetitive manual movement could raise different design questions; these
are illustrative task distinctions, not separate outcomes measured by the present score.
Similarly, changing the pace of work need not change who sets that pace. The operational
definition should therefore accompany any claim that an AI system addresses dull work.

The index also cannot directly measure boredom. That limitation follows from the items:
they describe occupational conditions rather than workers' subjective experience of
tedium. The component estimates neither establish nor remove that limitation.
DDD can still orient an inquiry, as \citet{ddd2026} propose, provided that researchers name
the particular condition, examine its local meaning, and ask workers whether it is one
they wish to change. This keeps the domain label useful without asking it to carry a
judgment that requires different evidence.

\subsection{Use profiles to formulate questions for contrasting settings}

Occupational profiles suggest one use for these distinctions: researchers can
look for settings that share a broad classification but differ in their condition profiles.
The comparison of construction laborers and housekeeping cleaners illustrates this use.
One profile scores highest in Dirty; the other scores highest in the combined Dull domain. Selecting both solely because their composite scores are relatively
high would obscure why each setting warrants attention. Inspecting the domains gives a
researcher a more specific basis for asking what work to observe and whom to involve.

This is a starting point for inquiry rather than an evaluated selection method. The next
step would be to examine whether the occupational description fits the workplaces and
workers actually recruited. Equipment, staffing, schedules, disabilities, experience, and
bargaining power can make work in the same occupation differ substantially. Workers may
also disagree about which responsibilities they wish to retain and which conditions they
want changed. Participatory studies demonstrate the value of eliciting those distinctions
\citep{welfare2019,spektor2023}.

A proposed system then needs a task-to-condition argument. For example, a dispatch
assistant might affect planning work without changing the physical hazards summarized in a
driver's occupational profile. A researcher would need to establish which part of the
work the assistant changes and whether that change reaches the claimed hazard or burden.
This is an illustrative evaluation question, not an intervention tested here. Keeping
the task and condition explicit prevents an occupation's exposure score from serving as
evidence that a particular system will improve its working conditions.

The supplementary composition estimates can inform attention to recruitment coverage,
but their selected CPS matches do not characterize every occupation in the group.
Likewise, the employment totals describe jobs represented in historical OEWS data, not
the number of people who want an intervention. These distinctions matter when moving from
a descriptive classification to a study involving actual workplaces and participants.

\subsection{Evaluate relief alongside redistribution and control}

Once a condition and proposed intervention have been specified, an evaluation should ask
whether that condition changes in use. It should also examine where the work goes.
Integration may introduce verification, repair, or coordination tasks, and time saved on
one task may become additional work elsewhere \citep{fox2023patchwork}. The relevant
comparison is therefore between work arrangements before and after the intervention,
including the supporting work that a task-level exposure rubric does not describe.

Control requires its own evidence. Standalone-model and additional-software exposure
describe technical requirements, not who chooses delegation or can override a system.
Either configuration could be used under worker discretion or managerial direction.
Labor process theory suggests examining who makes those decisions, who handles failures,
and who receives the gains from a change \citep{laborprocess2026}. These questions extend
the worker-centered critique of automation arguments \citep{rosie2025}; they are not
outcomes of the occupational regressions.

The exposure measures also set limits on technological claims. Low LLM exposure or AIOE
cannot establish protection from algorithmic management or other automation. Hospitality
research, for example, shows how algorithmic assignment can affect workload and control
\citep{spektor2023}. Researchers should state the technology and work arrangement they
evaluate, then compare the proposed intervention with organizational alternatives such as
staffing changes or safer equipment where appropriate. An occupational profile helps name
the condition under discussion; workers and workplace evidence are needed to judge the
change.

\section{Limitations and Scope}

\paragraph{Content and worker evidence.}
The index covers a selected subset of occupational conditions. It omits job security,
meaning, social relations, schedule instability, emotional demands, and other dimensions
of work. Equal domain weights and available-item scoring are analytic choices rather than
estimates of perceived severity. Protective equipment and outdoor work have context-dependent
meanings, and occupational averages conceal variation among workplaces and workers.
Worker-based content validation, test--retest evidence, and subjective boredom measurements
are absent. The index should be interpreted through its items rather than as a validated
measure of desirability.

\paragraph{Shared sources and incomplete comparisons.}
The condition ratings and both primary exposure measures use O*NET information. The hybrid
score also incorporates theoretical feasibility. Agreement across these measures therefore
does not eliminate common-source dependence. Adjustment covers a particular set of
occupational characteristics; human-rated LLM exposure and teleworkability comparisons
were not run. The fatality ratio combines mismatched periods and populations and provides
only a limited correspondence check, not aligned occupational-health validation.

\paragraph{Linkage and time periods.}
Detailed specialties receive equal weight within six-digit occupations, and item suppression
can change which specialties contribute. Employment comes from May 2021 and workforce
composition from 2022. The totals are historical, and the CPS title linkage leaves selected
subsets with unequal coverage. Exact title matching does not guarantee equivalent
classification boundaries. These choices, together with the exclusion of self-employment from OEWS employment counts,
limit generalization to the full US workforce and to other countries.

\paragraph{Inference and technological scope.}
The intervals rely on regression assumptions and a modest number of clusters. They exclude
uncertainty from source measurement, item selection, and linkage. Secondary comparisons
are exploratory; relative classification cutoffs have no
natural boundary. The exposure measures do not cover all robotics, embodied AI, mechanization,
or algorithmic management. Platform scores also depend on access, usage patterns, product
capabilities, and classification rules. The data contain no observed intervention that would
identify effects on burdens, autonomy, or displacement.

\section{Conclusion}

Across 754 occupations with employment data, higher LLM exposure and AIOE
are associated with lower scores on selected working conditions. The composite coefficients
remain negative across several scoring checks, but become smaller after physical demands
and other occupational characteristics enter the models. The fully adjusted LLM intervals
include zero, whereas the AIOE intervals remain below zero. Separating repetition from
restricted discretion reveals further distinctions hidden by the combined Dull score.
These findings show why an occupational exposure comparison needs explicit condition
definitions, component results, and a clear account of the occupational differences being
held constant. Occupational profiles can inform questions for HCI research; establishing
relief requires tracing a specific intervention to a condition workers want changed.

\section*{Research Transparency}

The analysis uses public aggregate occupational data and collected no new participant data.
Table~\ref{tab:sources} lists the data sources.

Generative AI tools assisted with analysis-code development, literature discovery, and
manuscript drafting and revision. I am responsible for the analytical decisions,
interpretations, references, and final text.

\bibliographystyle{ACM-Reference-Format}
\bibliography{refs}
\pagebreak
\appendix
\section{Item Specification and Linkage Details}
\label{app:items}

Table~\ref{tab:items} lists the primary domain assignments.
All items use the Work Context CX scale, which has item-specific response anchors. The two discretion items are reverse-scored.
The additional physical-demand items are 4.C.2.d.1.c (climbing), 4.C.2.d.1.e
(kneeling/\allowbreak crouching/\allowbreak stooping/\allowbreak crawling), 4.C.2.d.1.h (bending/twisting), and 4.C.2.d.1.b
(standing). Degree of Automation, 4.C.3.b.2, is excluded from the primary index because
it describes existing automation.

\begin{table}[h]
\caption{Items in the primary working-condition index. The table abbreviates source wording; exact item names and response anchors are available from O*NET. The two discretion items are reverse-scored.}
\label{tab:items}
\small
\begin{tabularx}{\linewidth}{llX}
\toprule
Domain & Element ID & Condition \\
\midrule
Dangerous & 4.C.2.c.1.a & Radiation \\
 & 4.C.2.c.1.b & Disease or infections \\
 & 4.C.2.c.1.c & High places \\
 & 4.C.2.c.1.d & Hazardous conditions \\
 & 4.C.2.c.1.e & Hazardous equipment \\
 & 4.C.2.c.1.f & Minor burns, cuts, bites, or stings \\
 & 4.C.2.e.1.d & Common protective or safety equipment \\
 & 4.C.2.e.1.e & Specialized protective or safety equipment \\
Dirty & 4.C.2.a.1.c & Outdoors, exposed to all weather \\
 & 4.C.2.a.1.d & Outdoors, under cover \\
 & 4.C.2.b.1.a & Distracting or uncomfortable noise \\
 & 4.C.2.b.1.b & Very hot or cold temperatures \\
 & 4.C.2.b.1.c & Extremely bright or inadequate lighting \\
 & 4.C.2.b.1.d & Contaminants \\
 & 4.C.2.b.1.e & Cramped workspace or awkward positions \\
 & 4.C.2.b.1.f & Whole-body vibration \\
Dull & 4.C.2.d.1.i & Time making repetitive motions \\
 & 4.C.3.b.7 & Importance of repeating the same tasks \\
 & 4.C.3.d.3 & Pace determined by equipment \\
 & 4.C.3.a.4 & Freedom to make decisions (reversed) \\
 & 4.C.3.b.8 & Determine tasks, priorities and goals (reversed) \\
\bottomrule
\end{tabularx}
\end{table}

\subsection{Processing and model details}
\label{app:linkage}

The LLM columns are \texttt{dv\_rating\_alpha}, \texttt{dv\_rating\_beta}, and
\texttt{dv\_rating\_gamma}. They appear in \texttt{occ\_level.csv}.
The \texttt{dv\_rating} prefix identifies GPT-4 ratings in the source repository.
AIOE comes from Appendix A of \path{AIOE_DataAppendix.xlsx}.
The classification crosswalk is O*NET's \path{2010_to_2018_SOC_Crosswalk.xlsx}.
I retained unique source--target pairs and used direct-code matching only for targets
without a crosswalked value.

At the six-digit level, Job Zone is the rounded mean of the detailed occupation ratings.
The computation uses nearest-even rounding for exact half values. The CR1 covariance
multiplier is $G/(G-1)\,(n-1)/(n-k)$, where $G$ is the number of major groups,
$n$ is the number of observations, and $k$ includes all fitted coefficients. The
reference distribution has $G-1$ degrees of freedom. Exposure standardization uses equal
occupation weights in the scored sample for the primary and adjustment models. Supplementary
component and hybrid models use the reference populations specified below.

\section{Measurement and Sample Diagnostics}
\label{app:diagnostics}

\subsection{Covariance among conditions}

Standardized-item alpha is 0.859 for Dangerous, based on
756 occupations with complete items in that domain. For Dirty,
alpha is 0.940 on 772 domain-complete occupations;
for Dull, it is 0.760 on 779. These sample sizes
describe the covariance calculations and differ from the regression sample sizes.
The coefficients summarize covariation among occupational ratings; they do not measure
the reliability of individual workers' evaluations or establish complete content coverage.

\subsection{Fatality-ratio correspondence}

The exploratory fatality ratio divides CFOI fatality counts from 2024 by OEWS employment
from 2021. The sources differ in population coverage as well as period, so the ratio is
not an official occupational fatality rate. Aligned populations and time at risk are
necessary for such rates \citep{galantucci2022}. Rank correlation does not resolve that
denominator mismatch.

Across 170 matched occupations, the Spearman correlation is
0.555 for Dangerous,
0.684 for Dirty,
0.570 for the composite, and
0.122 for Dull. Matching on exact titles gives
the same analytic overlap and correlations. Dirty's larger correlation is consistent
with physical content shared by the domains, whose direct correlation is reported in the
main text. The diagnostic provides limited correspondence with fatal injuries and cannot
validate subjective desirability or boredom.

\subsection{Primary coefficients and sample sensitivity}

Table~\ref{tab:primary} gives all primary coefficients and matching intervals.

\begin{table}[t]
\caption{Complete primary estimates: coefficient [95\% confidence interval], in condition-score standard deviations per AI-exposure standard deviation. All models include SOC major-group fixed effects and use 754 occupations and 22 clusters. Occ. denotes equal occupation weights; Emp. denotes employment weights. Intervals use CR1 covariance and a $t$ reference distribution.}
\label{tab:primary}
\centering\small
\begin{tabular}{llcc}
\toprule
Outcome & Weight & LLM exposure & AIOE \\
\midrule
Composite & Occ. & $-$0.350 [$-$0.470, $-$0.229] & $-$0.528 [$-$0.715, $-$0.341] \\
 & Emp. & $-$0.285 [$-$0.362, $-$0.208] & $-$0.425 [$-$0.517, $-$0.332] \\
Dangerous & Occ. & $-$0.322 [$-$0.460, $-$0.183] & $-$0.413 [$-$0.584, $-$0.243] \\
 & Emp. & $-$0.187 [$-$0.313, $-$0.061] & $-$0.300 [$-$0.439, $-$0.161] \\
Dirty & Occ. & $-$0.307 [$-$0.426, $-$0.188] & $-$0.425 [$-$0.611, $-$0.240] \\
 & Emp. & $-$0.202 [$-$0.305, $-$0.100] & $-$0.334 [$-$0.508, $-$0.159] \\
Dull & Occ. & $-$0.243 [$-$0.354, $-$0.131] & $-$0.478 [$-$0.671, $-$0.284] \\
 & Emp. & $-$0.320 [$-$0.522, $-$0.118] & $-$0.425 [$-$0.654, $-$0.195] \\
\bottomrule
\end{tabular}
\end{table}

Omitting major groups one at a time retains negative composite coefficients. For LLM
exposure, the range is $-$0.369 to
$-$0.320 under equal occupation weights and
$-$0.300 to $-$0.266
under employment weights. For AIOE, the range is $-$0.571
to $-$0.472 under equal occupation weights and
$-$0.447 to $-$0.407
under employment weights. Models with equal occupation weights using all scored
occupations also retain negative composite coefficients. These checks address particular
sample choices; they do not test omitted conditions or establish the meaning of the index.

\section{Workforce Composition in the Joint Groups}
\label{app:composition}

I linked the 2022 CPS composition table to official SOC occupation titles using exact
normalized-title matches and retained only unique source and target matches. Normalization
removed punctuation, footnote markers, case, and repeated whitespace. It did not establish
equivalence where Census and SOC categories differed in scope. I excluded unresolved
titles rather than assigning approximate matches and required all four reported composition
shares. I then averaged occupational shares using OEWS employment weights within the same
joint groups used in the main text. Race and Hispanic or Latino ethnicity overlap and
should not be summed.

The complete composition linkage contains 244 occupations and represents
72.1\% of primary-sample employment. Within the higher-score,
lower-exposure group, the linkage covers 75 occupations and
68.6\% of that group's employment. Within the contrast
group, it covers 111 occupations and
76.5\% of employment. Table~\ref{tab:composition} therefore
describes selected subsets with different coverage.

\begin{table}[t]
\caption{Workforce composition in the CPS-matched subsets. Entries are occupational workforce percentages averaged using OEWS employment weights. Both groups use the same joint classifications as the main analysis. Race and Hispanic ethnicity overlap.}
\label{tab:composition}
\begin{tabularx}{\linewidth}{Xrrrr}
\toprule
Set & Women & Black & Asian & Hispanic \\
\midrule
Higher score, lower exposure & 36.9 & 16.2 & 4.2 & 26.2 \\
Lower score, higher exposure & 56.1 & 10.9 & 7.7 & 13.5 \\
\bottomrule
\end{tabularx}
\end{table}

The matched higher-score, lower-exposure subset has larger Black and Hispanic shares and
smaller female and Asian shares than the matched contrast subset. This describes occupational
composition within the linkage. Differential coverage, CPS sampling variation, and the
different years of composition and employment weights limit inference beyond these subsets.
The comparison does not estimate individual AI use, preferences, or demographic effects
on exposure, and exposure itself does not identify a benefit or harm.

\section{Supplementary Exposure Measures}
\label{app:supplementary}

\subsection{Standalone-model and additional-software components}

The Eloundou rubric defines $\beta=\alpha+\tfrac12(\gamma-\alpha)$, where $\alpha$
describes standalone-model exposure and $\gamma-\alpha$ the additional exposure requiring
software built around the model \citep{eloundou2024}. I standardized the two components
with equal occupation weights within the primary employment sample and entered them in
the same model. Their coefficient difference and its interval use the covariance from
that joint fit.

Under equal occupation weights, the standalone coefficient minus the additional-software
coefficient is 0.208
[0.122, 0.293] for the composite and
0.342 [0.216, 0.467] for Dull.
This is a comparison of technical requirements. It does not identify
automation versus augmentation or estimate displacement under either arrangement.

\subsection{Hybrid exposure construction and linkage}

The supplementary source is the March 2026 measure of \citet{massenkoff2026}, which uses
platform traffic from August and November 2025. It combines theoretical feasibility,
classified work-related platform activity, and workflow weighting. The provider's appendix
specifies a usage threshold, a feasibility gate, and greater weight for automated workflows
\citep{massenkoff2026appendix}. These choices are already encoded in the released task
scores; the measure is not simply observed adoption.

I matched normalized task text to O*NET task statements and excluded normalized keys that
had conflicting source scores. I averaged the matched task-record scores within six-digit
occupations, retaining occupations with scores for at least 60\% of their task records.
Missing matches remained missing, while source zeros were retained. This average gives
equal weight to task records, including their occurrence across detailed specialties.
It differs from the provider's time-weighted occupational measure. I standardized the
hybrid score with equal occupation weights within its retained sample.

The linkage excluded 3 ambiguous normalized task keys and
matched 18,613 of 18,838 records
(98.8\%) in the full O*NET task file. That match rate precedes restrictions
to analytic occupations and the minimum-coverage rule. The retained occupational sample
contains 753 occupations. Its coverage relative to primary employment is
reported in Table~\ref{tab:samples}.

\subsection{Hybrid exposure results}

The employment-weighted mean hybrid score is 0.205 in the lowest
composite-score quartile and 0.012 in the highest. Quartile cutoffs give
equal weight to occupations. These values are task-score averages on the hybrid scale,
not percentages of workers using AI or shares of work time automated.

The composite coefficient for standardized hybrid exposure is
$-$0.248 [$-$0.314, $-$0.182] under equal occupation weights
and $-$0.110 [$-$0.231, 0.011] under employment weights.
Both point estimates are negative, but the employment-weighted interval includes zero.
The results extend the specification comparison with a differently constructed measure;
they are not an independent replication of adoption or relief because the score also
incorporates theoretical exposure.

\end{document}